\documentclass[conference]{IEEEtran}
\IEEEoverridecommandlockouts

\usepackage{cite}
\usepackage{amsmath,amssymb,amsfonts}
\usepackage{algorithmic}
\usepackage{graphicx}
\usepackage{textcomp}
\usepackage{xcolor}
\usepackage{multirow}
\usepackage[ruled,vlined]{algorithm2e}
\usepackage{url}
\def\BibTeX{{\rm B\kern-.05em{\sc i\kern-.025em b}\kern-.08em
    T\kern-.1667em\lower.7ex\hbox{E}\kern-.125emX}}
\begin{document}

\title{Med-2E3: A 2D-Enhanced 3D Medical Multimodal Large Language Model}

\renewcommand{\thefootnote}{\fnsymbol{footnote}}
\author{
Yiming Shi\thanks{*~Equal contribution}\textsuperscript{1*} \quad Xun Zhu\textsuperscript{1*} \quad Kaiwen Wang\textsuperscript{1} \quad Ying Hu\textsuperscript{1} \quad Chenyi Guo\textsuperscript{1} \quad Miao Li\thanks{\dag~Corresponding authors: Ji Wu and Miao Li}\textsuperscript{1\dag} \quad Ji Wu\textsuperscript{1, 2, 3\dag}\\
\textsuperscript{1} Department of Electronic Engineering, Tsinghua University\\
\textsuperscript{2} College of AI, Tsinghua University\\
\textsuperscript{3} Beijing National Research Center for Information Science and Technology \\
{\tt\small \{sym23, zhu-x24, wkw23, yinghu\_yh\}@mails.tsinghua.edu.cn}\\
{\tt\small \{guochy, miao-li, wuji\_ee\}@tsinghua.edu.cn}
}

\maketitle

\begin{abstract}
3D medical image analysis is essential for modern healthcare, yet traditional task-specific models are inadequate due to limited generalizability across diverse clinical scenarios. Multimodal large language models (MLLMs) offer a promising solution to these challenges. However, existing MLLMs have limitations in fully leveraging the rich, hierarchical information embedded in 3D medical images. Inspired by clinical practice, where radiologists focus on both 3D spatial structure and 2D planar content, we propose Med-2E3, a 3D medical MLLM that integrates a dual 3D-2D encoder architecture. To aggregate 2D features effectively, we design a Text-Guided Inter-Slice (TG-IS) scoring module, which scores the attention of each 2D slice based on slice contents and task instructions. To the best of our knowledge, Med-2E3 is the first MLLM to integrate both 3D and 2D features for 3D medical image analysis. Experiments on large-scale, open-source 3D medical multimodal datasets demonstrate that TG-IS exhibits task-specific attention distribution and significantly outperforms current state-of-the-art models. The code is available at: \url{https://github.com/MSIIP/Med-2E3}
\end{abstract}

\begin{IEEEkeywords}
3D medical multimodal large language models, 3D medical image analysis, 3D medical report generation, 3D medical visual question answering
\end{IEEEkeywords}

\section{Introduction}
3D medical images play a pivotal role in modern healthcare, providing critical 3D structural information\cite{app3d}. However, conventional models for 3D medical image analysis, whether based on traditional image processing techniques or early deep learning, are primarily designed for specific, single-modality tasks\cite{app3d}. Their limited capability in handling complex multi-modal tasks, such as medical report generation and visual question answering (VQA) \cite{rgvqa}, impedes the development of medical AI in real-world clinical scenarios. The recent success of large language models (LLMs) like ChatGPT \cite{gpt4} has catalyzed the development of multimodal LLMs (MLLMs) \cite{blip2,flamingo,qwenvl,internvl,tinyllava,tinyllavafactory}. Driven by this wave, researchers have actively pursued the development of medical MLLMs \cite{llavamed,medflamingo,medpalmm,gmai}.

However, most current medical MLLMs \cite{llavamed,medflamingo,medpalmm,gmai} still focus on 2D medical image analysis, and their architectures do not inherently support 3D medical image inputs. Some researchers have begun to extend the 2D MLLM paradigm to the 3D domain \cite{medblip,m3d,ctrate,3dctgpt,diallama,braingpt}. For example, Bai et al. \cite{m3d} constructed M3D-Data, a large-scale multi-modal dataset of whole-body CT scans, and pretrained a 3D vision encoder, M3D-CLIP, upon which they built a 3D medical MLLM, M3D-LaMed. Similarly, Hamamci et al. \cite{ctrate} built CT-RATE, a multi-modal dataset for non-contrast chest CTs. Based on CT-RATE, they trained a 3D encoder, CT-CLIP, and a 3D medical MLLM, CT-CHAT.

\begin{figure}[t]
\centerline{\includegraphics[width=0.9\linewidth]{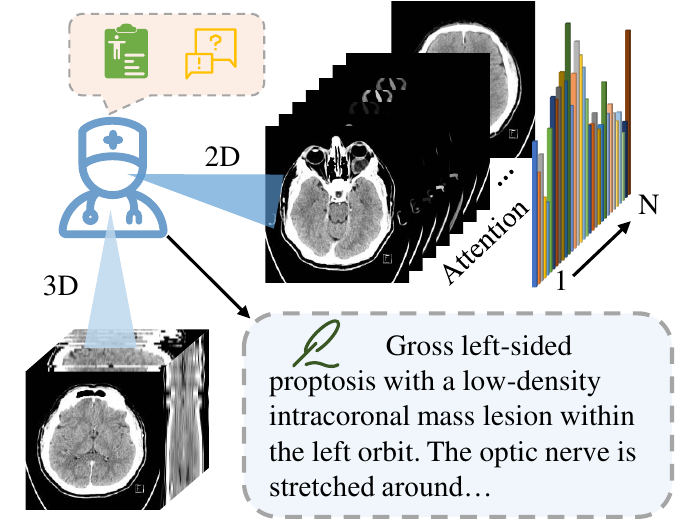}}
\caption{Diagnostic workflow of interpreting 3D medical images. Radiologists typically analyze 3D medical images from both global (3D) and local (2D) perspectives, enabling them to focus on spatial structures and planar content. They allocate attention differently to slices based on their content and the specific task requirements.}
\label{fig1}
\end{figure}

\begin{figure*}[t]
\centerline{\includegraphics[width=0.9\linewidth]{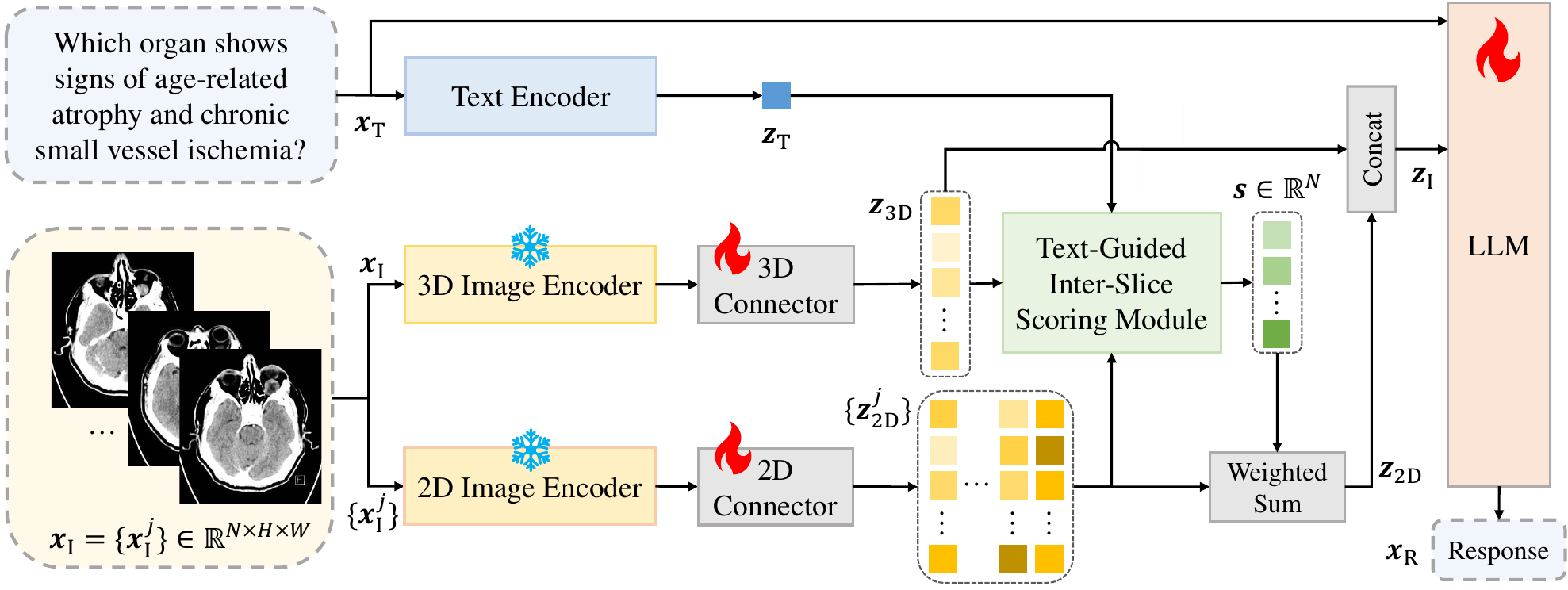}}
\caption{Overall framework of Med-2E3. Med-2E3 employs a dual 3D-2D encoder architecture to extract complementary features ($\mathbf{z}_{\text{3D}}$ and $\{\mathbf{z}_{\text{2D}}^j\}$), respectively. Attention scores $\mathbf{s}\in\mathbb{R}^N$, calculated by the TG-IS scoring module, are used to aggregate 2D features $\{\mathbf{z}_{\text{2D}}^j\}$. Aggregated 2D features $\mathbf{z}_{\text{2D}}$ are concatenated with 3D features $\mathbf{z}_{\text{3D}}$. Finally, LLM process the 2D-enhanced 3D medical image features $\mathbf{z}_{\text{I}}$ with text input $\mathbf{x}_{\text{T}}$ to generate the response $\mathbf{x}_{\text{R}}$.}
\label{fig:framework}
\end{figure*}

While M3D-LaMed and CT-CHAT represent important explorations, they predominantly adopt a straightforward strategy of replacing the 2D vision encoder with a 3D counterpart. This approach, however, fails to fully explore the diagnostic workflow of interpreting 3D medical images in clinical practice. As shown in Fig.~\ref{fig1}, radiologists typically analyze 3D images from a dual perspective, integrating global and local views: they grasp the spatial structures from a 3D perspective while capturing planar details from 2D slices. Compared to mature 2D encoders \cite{clip,siglip,siglip2}, pre-training resources for 3D encoders are considerably scarcer. As a result, while pre-trained 3D encoders excel at capturing spatial structures, their ability to perceive local details is compromised.

Inspired by this clinical insight, we enhance the LLaVA architecture by introducing a dual 3D-2D encoder architecture to synergistically capture both global structural and local detailed features from 3D medical images. Furthermore, to simulate how radiologists dynamically allocate their attention across different slices based on the task, we design a text-guided inter-slice (TG-IS) scoring module to adaptively compute attention distribution.

In summary, our main contributions are as follows:

\begin{itemize}
\item  We propose \textbf{Med-2E3}, which is the first 3D medical MLLM to integrate a dual 3D-2D encoder architecture.
\item We design the \textbf{TG-IS} scoring module, which simulates the task-dependent inter-slice attention mechanism of radiologists during radiological reviews.
\item Med-2E3 achieves \textbf{SOTA} performance on both medical report generation and medical VQA tasks, demonstrating the superiority of our proposed approach.
\end{itemize}

\section{Method}

This section presents the overall framework of our proposed Med-2E3. As shown in Fig.~\ref{fig:framework}, the input to Med-2E3 consists of a 3D medical image $\mathbf{x}_{\text{I}}$ and corresponding task instructions $\mathbf{x}_{\text{T}}$, with the output being a textual response $\mathbf{x}_{\text{R}}$. Med-2E3 can be applied to diverse clinical scenarios, including 3D medical report generation and 3D medical VQA tasks.

First, as described in Section \ref{sec:extract}, Med-2E3 employs 3D, 2D, and text encoders to extract preliminary features from the 3D image and text. Then, as detailed in Section \ref{sec:scoring}, a text-guided inter-slice scoring module is designed to calculate the attention scores for each slice based on these preliminary features. Subsequently, the 2D features are aggregated based on the attention scores and then concatenated with the preliminary 3D features. LLM process the 2D-enhanced 3D features with the text input to generate a response, as described in Section \ref{sec:enhance}. The loss function of Med-2E3 is defined in Section \ref{sec:loss}.

\subsection{Image and Text Feature Extraction}
\label{sec:extract}
We follow the common approach of extracting features using corresponding modality encoders for the input 3D image $\mathbf{x}_{\text{I}}$ and task instructions $\mathbf{x}_{\text{T}}$. It is important to note that the features extracted at this stage are preliminary. For example, the 3D image features will undergo specially designed enhancement operations in subsequent stages.

For task instructions in textual form, a text encoder is used to extract text features $\mathbf{z}_{\text{T}}$:
\begin{equation}
\mathbf{z}_{\text{T}}=f_{\text{T}}(\mathbf{x}_{\text{T}}).
\label{eq:text}
\end{equation}
Here, $f_{\text{T}}$ denotes the text encoder.

As illustrated in Fig.~\ref{fig:framework}, during image feature extraction, the 3D medical image is represented in two forms: a raw form $\mathbf{x}_{\text{I}}\in\mathbb{R}^{N\times H\times W}$ and a slice-based form $\{\mathbf{x}_{\text{I}}^j\in\mathbb{R}^{H\times W}\}$. The 3D and 2D feature extraction branches independently process these forms, producing complementary features $\mathbf{z}_{\text{3D}}$ and $\{\mathbf{z}_{\text{2D}}^j\}$, respectively:
\begin{equation}
\mathbf{z}_{\text{3D}} = f_{\text{I}}^{(\text{3D})}(\mathbf{x}_{\text{I}}),
\label{eq:3D}
\end{equation}
\begin{equation}
\mathbf{z}_{\text{2D}}^j = f_{\text{I}}^{(\text{2D})}(\mathbf{x}_{\text{I}}^j).
\label{eq:2D}
\end{equation}
Here, $f_{\text{I}}^{(\text{3D})}$ and $f_{\text{I}}^{(\text{2D})}$ denote the 3D and 2D feature extraction branches. The 3D feature extraction branch includes a 3D image encoder and a 3D connector, while the 2D feature extraction branch consists of a 2D image encoder and a 2D connector. Superscript $j$ represents the slice index, which ranges from $1$ to $N$.

Since both $f_{\text{I}}^{(\text{3D})}$ and $f_{\text{I}}^{(\text{2D})}$ utilize ViT-based \cite{vit} vision encoders and employ MLP-based connectors that modify only the hidden dimensionality without altering the tensor shape, the resulting representations $\mathbf{z}_{\text{3D}}$ and $\mathbf{z}_{\text{2D}}^j$ are one-dimensional sequences with lengths $L_1$ and $L_2$, respectively, and a hidden dimension of size $D$.

\subsection{Text-Guided Inter-Slice Scoring}
\label{sec:scoring}

To mimic the attention distribution mechanism used by radiologists, we design a text-guided inter-slice (TG-IS) scoring module to score the attention of each slice based on its contents and task instructions.

\begin{figure}[t]
\centerline{\includegraphics[width=0.9\linewidth]{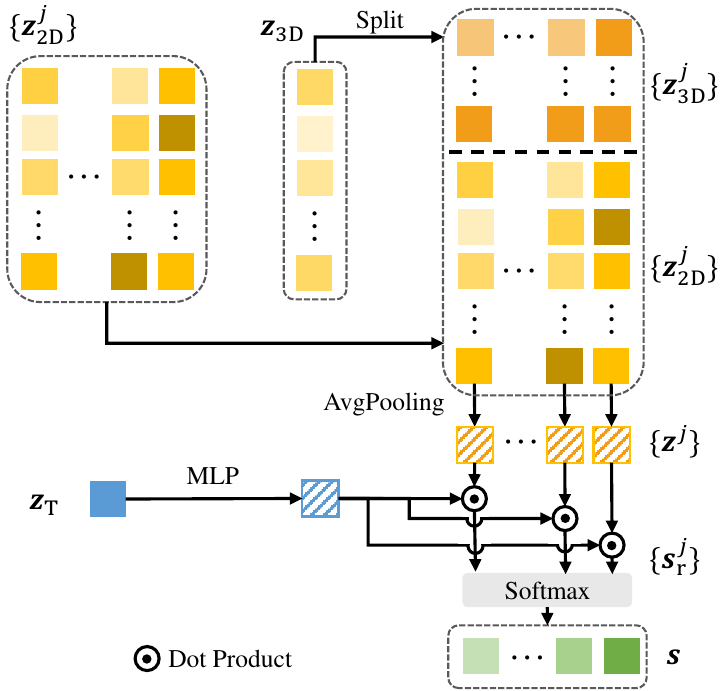}}
\caption{TG-IS scoring module. 3D features $\mathbf{z}_{\text{3D}}$ and 2D features $\{\mathbf{z}_{\text{2D}}^j\}$ are fused to derive the key features $\mathbf{z}^j$ for each slice. Attention scores $\mathbf{s} \in \mathbb{R}^N$ are computed by calculating the correlation $\{\mathbf{s}_{\text{r}}^j\}$ with the task instruction feature $\mathbf{z}_{\text{T}}$, followed by the softmax function for normalization.}
\label{fig:scoring}
\end{figure}

First, as shown in Fig.~\ref{fig:scoring}, based on the position of each slice, TG-IS module splits 3D features and selects relevant local features. Specifically, the serialized 1D features $\mathbf{z}_{\text{3D}} \in \mathbb{R}^{L_1\times D}$ are reshaped into a 3D form and the features corresponding to slice $j$ are selected as the local features $\mathbf{z}_{\text{3D}}^j \in \mathbb{R}^{L\times D}$.

Next, the features of each slice from the 3D and 2D representations are concatenated and averaged to form slice features:
\begin{equation}
\mathbf{z}^j = \text{AvgPooling}([\mathbf{z}_{\text{3D}}^j; \mathbf{z}_{\text{2D}}^j]) \in \mathbb{R}^D.
\label{eq:slicefeat}
\end{equation}
Here, $D$ denotes the feature dimension.

Since radiologists allocate varying levels of attention based on different tasks, the TG-IS scoring module is designed to mimic this task-specific attention distribution. In practice, the relevance of each slice $s_{\text{r}}^j$ to the current task is computed by calculating the dot product between each slice's features $\textbf{z}^j$ and the textual features, corresponding to the attention a radiologist would assign to each slice under the given task:
\begin{equation}
s^j_{\text{r}} = \text{MLP}(\mathbf{z}_{\text{T}}) \cdot \mathbf{z}^j.
\label{eq:scorer}
\end{equation}
The softmax function is then applied to normalize the relevance coefficients $\mathbf{s}_{\text{r}}\in\mathbb{R}^N$:
\begin{equation}
\mathbf{s} = \text{Softmax}(\mathbf{s}_r) \in \mathbb{R}^N,
\label{eq:score}
\end{equation}
to obtain the final attention score for each slice.

\subsection{2D-Enhanced 3D Features}
\label{sec:enhance}
During 3D feature enhancement, to reduce computational load, the 2D features $\{\mathbf{z}_{\text{2D}}^j\}$ extracted by the 2D branch are first aggregated based on the attention scores $\mathbf{s}\in\mathbb{R}^N$:
\begin{equation}
\mathbf{z}_{\text{2D}} = \sum_{j=1}^{N} s^j \mathbf{z}_{\text{2D}}^j.
\label{eq:2Dsum}
\end{equation}
Next, the 3D and aggregated 2D features are concatenated to form the final features of the 3D medical image:
\begin{equation}
\mathbf{z}_{\text{I}} = [\mathbf{z}_{\text{3D}}; \mathbf{z}_{\text{2D}}].
\label{eq:imgfeat}
\end{equation}

Finally, Med-2E3 inputs the 2D-enhanced 3D features along with the text features into the LLM, which processes them to generate the final textual response:
\begin{equation}
\mathbf{x}_{\text{R}} = f_{\text{LLM}}([\mathbf{z}_{\text{I}}; f_{\text{embed}}(\mathbf{x}_{\text{T}})]).
\label{eq:wholeModel}
\end{equation}

\subsection{Loss Function of Med-2E3}
\label{sec:loss}

In summary, the end-to-end processing function of Med-2E3 is defined as follows:
\begin{equation}
\mathbf{x}_{\text{R}} = f_{\text{Med-2E3}}(\mathbf{x}_{\text{I}}, \mathbf{x}_{\text{T}}),
\label{eq:overall}
\end{equation}
which is similar to the common MLLMs \cite{llava}. The corresponding loss function employed during training is defined as:
\begin{equation}
\mathcal{L}_{\text{Med-2E3}} = -\sum_{i=1}^{L} \log p_\theta(\mathbf{x}_{\text{R}}^{i} \,|\, \mathbf{x}_{\text{R}}^{<i}, \mathbf{x}_{\text{I}}, \mathbf{x}_{\text{T}})
\label{eq:loss}
\end{equation}

\section{Experiment}

This section presents a series of quantitative and qualitative experiments to systematically validate the effectiveness of our proposed Med-2E3, including the dual 3D-2D encoder architecture and TG-IS scoring module.

\begin{table*}[t]
\caption{Med-2E3 achieves state-of-the-art performance on medical report generation and VQA tasks of CT-RATE.}
\begin{center}
\begin{tabular}{c|ccc|ccc|ccc|c}
\hline
\multirow{2}[4]{*}{Method} & \multicolumn{3}{|c}{Report Generation} & \multicolumn{3}{|c}{Long Answer} & \multicolumn{3}{|c}{Short Answer} & \multicolumn{1}{|c}{Choice} \\
\cline{2-11}
& BLEU & ROUGE & METEOR & BLEU & ROUGE & METEOR & BLEU & ROUGE & METEOR & Accuracy \\
\hline
CT-CHAT-7B \cite{ctrate} & 38.88 & 32.71 & 21.65 & 47.02 & 48.45 & 28.20 & 27.46 & 45.59 & 15.47 & 83.73\\
CT-CHAT-8B \cite{ctrate} & 38.09 & 33.44 & 21.73 & 48.01 & 51.22 & 29.36 & 27.96 & 59.78 & 16.00 & 84.19 \\
CT-CHAT-70B \cite{ctrate} & 39.52 & 32.12 & 21.85 & 48.24 & 51.49& 29.48 & 27.44 & 59.19 & 15.53 & 84.35\\
\hline
Med-2E3 & \textbf{44.09} & \textbf{38.86} & \textbf{25.73} & \textbf{51.83} & \textbf{55.25} & \textbf{34.61} & \textbf{58.35} & \textbf{60.06} & \textbf{33.69} & \textbf{90.03} \\
\hline
\end{tabular}
\label{tab:sota-ctrate}
\end{center}
\end{table*}

\begin{table*}[t]
\caption{Med-2E3 achieves state-of-the-art performance on medical report generation and VQA tasks of M3D-Data.}
\begin{center}
\begin{tabular}{c|cccc|cccccc}
\hline
\multirow{2}[4]{*}{Method} & \multicolumn{4}{|c|}{Report Generation} & \multicolumn{5}{c}{Medical VQA} \\
\cline{2-10}
& BLEU & ROUGE & METEOR & BERTScore & BLEU & ROUGE & METEOR & BERTScore & Accuracy \\
\hline
RadFM \cite{radfm} & 12.23 & 16.49 & 11.57 & 87.93 & 16.39 & 26.13 & 21.33 & 88.72 & 19.79 \\
M3D-LaMed \cite{m3d} & 37.30 & 40.70 & 36.74 & 88.03 & 53.63 & 57.23 & 37.65 & 92.38 & 80.16 \\
\hline
Med-2E3 & \textbf{55.03} & \textbf{57.85} & \textbf{54.55} & \textbf{91.50} & \textbf{58.68} & \textbf{62.21} & \textbf{41.65} & \textbf{93.27} & \textbf{82.60} \\
\hline
\end{tabular}
\label{tab:sota-m3d}
\end{center}
\end{table*}

\subsection{Datasets and Evaluation Metrics}
\label{sec:data}
Our experiments are conducted on two large-scale 3D medical multimodal datasets: CT-RATE \cite{ctrate} and M3D-Data \cite{m3d}. Although CT-RATE focuses on non-contrast chest CT and M3D-Data covers whole-body CT, both provide data for medical report generation and VQA, supporting our comprehensive evaluation. Regarding medical VQA task categorization, CT-RATE divides its questions into three types: long-answer, short-answer, and multi-choice. M3D-Data primarily classifies its VQA tasks into open-ended and closed-ended.

To balance experimental efficiency and comprehensive validation, the experiments in section \ref{sec:dual} and \ref{sec:tgis} were conducted on a sampled subset of the training data, while the official test splits were kept intact. Specifically, for both CT-RATE and M3D-Data, we sampled 10K instances for the report generation task. For VQA, we sampled 10K instances from each category, resulting in 30K samples from CT-RATE, and 100K samples from M3D-Data. During all experiments, we strictly follow a separate training and testing paradigm. The training sets of CT-RATE and M3D-Data are never mixed, and models are trained and evaluated on their respective datasets.

For evaluation metrics, we adhere to the established standards set by CT-RATE \cite{ctrate} and M3D-Data \cite{m3d}. For multiple-choice VQA tasks, we use Accuracy as the evaluation metric. For report generation and other open-ended VQA tasks, we employ natural language generation metrics, including BLEU, ROUGE, METEOR, and BERTScore.

\subsection{Implementation and Training Details}
\label{sec:model}

To ensure fair comparison and manage computational resources, our model implementation largely follows the setup of M3D-LaMed \cite{m3d}. Specifically, we employ M3D-CLIP as the 3D vision encoder, SigLIP-256 \cite{siglip} as the 2D vision encoder, and Phi-3-mini as the LLM. All input CT volumes are resized to a uniform resolution of 32×256×256. For preprocessing, Hounsfield Unit (HU) values in CT-RATE data are clipped to the [-1000, 1000] range and then directly converted to grayscale images. The M3D-Data images are already in grayscale and require no additional processing.

Med-2E3 is trained using a two-stage strategy: pre-training followed by instruction fine-tuning. In the pre-training stage, we use the report generation data from either CT-RATE or M3D-Data and exclusively update the parameters of the connector module, keeping the rest of the model frozen. In the instruction fine-tuning stage, we use a mixture of medical report and VQA data and jointly fine-tune the 2D encoder, the connector, and the LLM.

The entire training process is conducted on two NVIDIA A800 (80G) GPUs. We utilize the ZeRO-3 distributed training strategy and bfloat16 mixed-precision to optimize memory usage and training speed. During the pre-training stage, the batch size is set to 32 with a gradient accumulation step of 1. In the instruction tuning stage, the batch size is set to 16, also with a gradient accumulation step of 1. For training on the CT-RATE dataset, following the settings in the CT-RATE paper \cite{ctrate}, we perform 1 epoch of pretraining and 1 epoch of instruction tuning. For training on the M3D-Data dataset, we follow the configuration in the M3D paper \cite{m3d}, conducting 3 epochs of pretraining and 4 epochs of instruction tuning.

\subsection{State-of-the-art Performance on Public Benchmarks}
\label{sec:sota}

As shown in Table \ref{tab:sota-ctrate}, Med-2E3 outperforms the previous state-of-the-art model, CT-CHAT, across both report generation and various VQA tasks on the CT-RATE dataset. Likewise, Table \ref{tab:sota-m3d} demonstrates that Med-2E3 surpasses the prior best model, M3D-LaMed, in both report generation and VQA tasks on the M3D-Data dataset. These consistent improvements across datasets and task types strongly validate the effectiveness of the Med-2E3 architecture.

\begin{table}[t]
\caption{Comparison between the dual 3D–2D encoder and single-encoder architecture.}
\begin{center}
\begin{tabular}{c|cccc|cc}
\hline
\multirow{2}[4]{*}{Method} & \multicolumn{4}{|c}{CT-RATE} & \multicolumn{2}{|c}{M3D-Data} \\
\cline{2-7}
& RG & Long & Short & Choice & RG & VQA \\
\hline
3D Only & 26.44 & 42.62 & 53.94 & 83.60 & 44.97 & 70.13 \\
2D Only & 24.06 & 41.33 & 52.50 & 83.51 & 36.96 & 71.71 \\
Med-2E3 & \textbf{29.96} & \textbf{44.80} & \textbf{54.20} & \textbf{85.36} & \textbf{53.21} & \textbf{72.11} \\
\hline
\end{tabular}
\label{tab:dual}
\end{center}
\end{table}

\begin{table}[t]
\caption{Comparison of different attention distribution.}
\begin{center}
\begin{tabular}{c|cccc|cc}
\hline
\multirow{2}[4]{*}{Method} & \multicolumn{4}{|c}{CT-RATE} & \multicolumn{2}{|c}{M3D-Data} \\
\cline{2-7}
& RG & Long & Short & Choice & RG & VQA \\
\hline
MaxPooling & 27.36 & 42.51 & 53.58 & 83.38 & 45.91 & 71.51 \\
AvgPooling & 28.56 & 43.77 & 54.05 & 84.58 & 48.50 & 71.47 \\
Gaussian & 27.09 & 42.66 & 53.24 & 83.12 & 46.42 & 71.33 \\
Random & 27.93 & 42.77 & 53.49 & 83.60 & 47.24 & 71.26 \\
Med-2E3 & \textbf{29.96} & \textbf{44.80} & \textbf{54.20} & \textbf{85.36} & \textbf{53.21} & \textbf{72.11} \\
\hline
\end{tabular}
\label{tab:tgis}
\end{center}
\end{table}

\begin{figure*}[t]
\centerline{\includegraphics[width=1\linewidth]{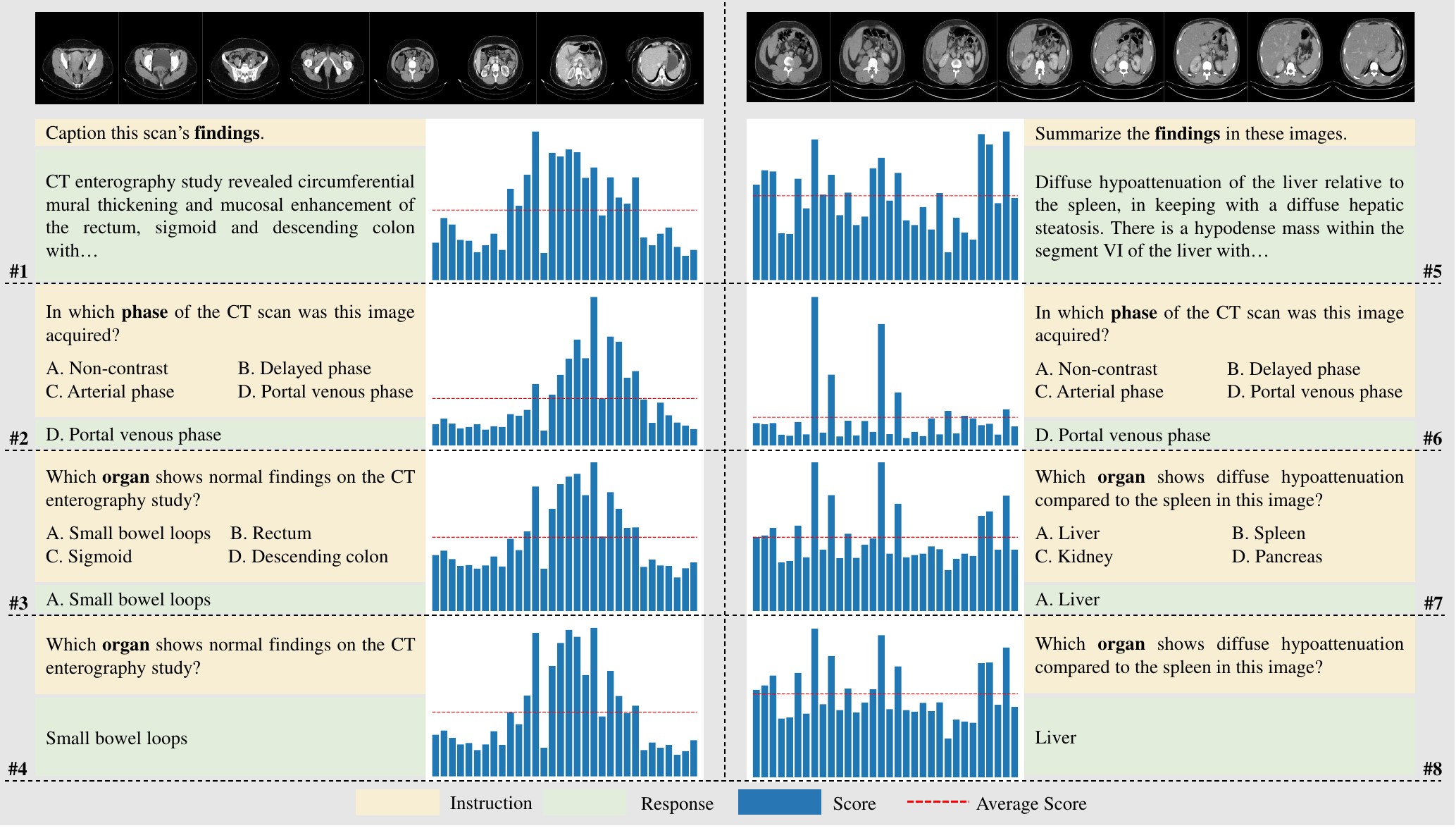}}
\caption{Visualization of two CT volumes and their corresponding question-answer pairs. To improve clarity, 8 slices were uniformly sampled from the full CT sequence. Questions 1 and 5 correspond to report generation tasks, while questions 2, 3, 6, 7 are closed-ended VQA tasks (multiple-choice), and questions 4 and 8 are open-ended VQA tasks (free-form). Keywords in each question are highlighted in \textbf{bold}. The \textcolor{red}{red} dashed line in the score distribution represents the average attention score ($1/32$). Distinct attention distributions are observed for different questions on the same CT volume, while similar distribution patterns emerge for the same question posed in different formats (multiple-choice and free-form). This demonstrates that Med-2E3 effectively captures essential task-relevant information.}
\label{fig6}
\end{figure*}

\subsection{Effectiveness of Dual 3D-2D Encoder Architecture}
\label{sec:dual}

To evaluate the effectiveness of the dual 3D–2D encoder architecture, we conducted experiments on the sampled CT-RATE and M3D-Data datasets, comparing the performance of single-encoder baselines with that of Med-2E3 architecture. The single-encoder baselines include: 3D only, which extracts 3D medical image features using M3D-CLIP; and 2D only, which extracts features from 3D images using SigLIP-256.

Evaluation results are presented in Table \ref{tab:dual}. Specifically, "RG" refers to the report generation task, "Long" to long-answer VQA, "Short" to short-answer VQA, and "Choice" of M3D-Data to closed-ended VQA. For report generation, long-answer and short-answer VQA tasks, BLEU scores are reported, while Accuracy is used for the Choice-related tasks. As illustrated in Table \ref{tab:dual}, benefiting from complementary feature representations obtained during the 3D medical image encoding stage, the dual 3D-2D encoder architecture of Med-2E3 consistently outperforms the single-encoder counterparts across different tasks and datasets.

\subsection{Effectiveness of Text-guided Inter-slice Module}
\label{sec:tgis}

To validate the effectiveness of the TG-IS scoring module, we designed and compared several alternative attention distribution strategies. Specifically, MaxPooling applies a max pooling operation across the slice channel dimension. AvgPooling performs average pooling across slices, resulting in equal attention scores for all slices. Gaussian assigns attention scores following a Gaussian distribution with zero mean and unit variance. Random distributes attention scores randomly across slices. All attention distributions are normalized to sum to one prior to being used for feature weighting.

The abbreviations and metrics in Table \ref{tab:tgis} are consistent with those defined in Section \ref{sec:dual}. As shown in in Table \ref{tab:tgis}, Med-2E3, which uses the TG-IS scoring module for feature aggregation, achieves the best performance across all attention distribution strategies. Among the alternative strategies, AvgPooling performs relatively well and serves as a competitive substitute. Although it may seem intuitive that attention distribution in 3D medical image analysis should follow a Gaussian pattern, experimental results suggest otherwise. This indicates that the attention distribution of radiologists is inherently more complex and must be adaptively adjusted based on the characteristics of the 3D medical image and specific task requirements.

\subsection{Case Study}
\label{sec:case}

To illustrate how different tasks yield distinct attention distributions, two representative CT volumes and four corresponding question-answer pairs are selected from M3D-Data, as shown in Fig.~\ref{fig6}. The results indicate that for the same question, different samples produce varying attention distributions, highlighting the influence of 3D image features during scoring.

For the same sample with different questions, the attention scores vary, as observed in questions \#2 and \#3 (also \#6 and \#7) of Fig.~\ref{fig6}, reflecting the influence of text guidance in the scoring process. A comparison between questions \#2 and \#3 (also \#6 and \#7) reveals that phase-related questions exhibit more extreme attention scores than organ-related questions. This trend suggests that phase-related questions require finer detail recognition, leading Med-2E3 to base its judgments on a smaller subset of slices.

Med-2E3 also demonstrates an ability to focus on the essence of the questions. As shown in Fig.~\ref{fig6}, comparing questions \#3 and \#4 (also \#7 and \#8), which differ in format (multi-choice vs. free-form) but share identical content, reveals that Med-2E3 generates similar attention distributions for both. This indicates that the model prioritizes essential information when extracting features from the task instruction. These findings suggest that Med-2E3 exhibits a certain level of reliability and interpretability.

\section{Conclusion}

In summary, we propose Med-2E3, the first MLLM to integrate both 3D and 2D encoders for 3D medical image analysis. To aggregate 2D features effectively, a TG-IS scoring module is introduced, computing attention scores for different slices based on slice content and task instructions. Med-2E3 achieves SOTA performance on CT-RATE and M3D-Data.

Limited by data availability, our experiments are mainly conducted on CT volumes and relatively simple VQA tasks. We anticipate future releases of multimodal 3D medical datasets covering MRI, PET, and more complex reasoning-based VQA tasks. Med-2E3 is expected to generalize well to broader 3D modalities and more challenging VQA scenarios.

We would like to emphasize that, beyond achieving excellent performance, the proposed TG-IS scoring module enhances the interpretability of the model’s decision-making process, contributing to greater transparency in clinical practice. Moreover, we aim to provide new insights for the 3D medical image analysis research community by integrating 3D and 2D encoders to leverage complementary features.

\section*{Acknowledgment}

This study was funded by Beijing Natural Science Foundation NO.L251072 and Beijing Natural Science Foundation NO.4252046.

\bibliographystyle{IEEEtran}
\bibliography{IEEEabrv,main}

\clearpage
\appendix

\subsection{3D Feature Transformation}
In the TG-IS scoring module, complete slice features $\{\mathbf{z}^j\}$ are formed by dividing and reorganizing the 3D features $\mathbf{z}_{\text{3D}}$ into a multi-slice format $\{\mathbf{z}_{\text{3D}}^j\in\mathbb{R}^{L\times D}\}$, which are then concatenated with their corresponding 2D features $\{\mathbf{z}_{\text{2D}}^j\}$.

Assuming the original image size is $N\times H\times W$, the features extracted by the 3D encoder with a patch size of $N_1\times H_1\times W_1$, are reshaped into a 1D vector of length $L_1$:
\begin{equation}
L_1 = \frac{N}{N_1}\cdot\frac{H}{H_1}\cdot\frac{W}{W_1}.
\label{eq:L1}
\end{equation}
Subsequently, the features are processed through a 3D connector with a pooling layer. Assuming the pooling operation downsamples by a factor of $P$, the resulting 3D features are further reduced to a 1D vector of length $L_2$:
\begin{equation}
L_2=\frac{L_1}{P^3}=\frac{N}{N_1\cdot P}\cdot\frac{H}{H_1\cdot P}\cdot\frac{W}{W_1\cdot P}.
\label{eq:L2}
\end{equation}
Consequently, the final 3D features $\mathbf{z}_{\text{3D}}$ have a shape of $L_2\times D$.

As shown in Algorithm \ref{alg:tg-is}, the one-dimension 3D features $\mathbf{z}_{\text{3D}}$ are initially reshaped into a 3D format, resulting in the following feature shape:
\begin{equation}
\frac{N}{N_1\cdot P}\times\frac{H}{H_1\cdot P}\times\frac{W}{W_1\cdot P}\times D.
\label{eq:shape1}
\end{equation}
To align the first dimension with the number of 2D features $\{\mathbf{z}_{\text{2D}}^j\}$, each layer of the 3D features is replicated $N_1\cdot P$ times. Rather than directly concatenating the replicated layers, an \textbf{interleaved concatenation} strategy is applied, resulting in the following feature shape:
\begin{equation}
N\times\frac{H}{H_1\cdot P}\times\frac{W}{W_1\cdot P}\times D.
\label{shape2}
\end{equation}
Finally, the features are reshaped into $N$ instances of $\mathbf{z}_{\text{3D}}^j$, each with a shape of $L\times D$, where $L$ is defined as:
\begin{equation}
L=\frac{H}{H_1\cdot P}\times\frac{W}{W_1\cdot P}.
\label{eq:L}
\end{equation}

\begin{algorithm}
\SetAlgoLined
\caption{Transformation of 3D features for alignment with 2D features in the first dimension.}
\label{alg:tg-is}

\KwIn{3D features $\mathbf{z}_{\text{3D}}$ of shape $(L_2, D)$, where \\
\[
L_2 = \frac{N}{N_1 \cdot P} \cdot \frac{H}{H_1 \cdot P} \cdot \frac{W}{W_1 \cdot P}
\]}
\KwOut{Transformed 3D features $\{z_{\text{3D}}^j\}$ of shape $(N, L, D)$, where \\
\[
L = \frac{H}{H_1 \cdot P} \cdot \frac{W}{W_1 \cdot P}
\]
}

\textbf{Step 1: Reshape} the input $\mathbf{z}_{\text{3D}}$ into a 3D format:
\[
\text{New shape} \gets \left( \frac{N}{N_1 \cdot P}, \frac{H}{H_1 \cdot P}, \frac{W}{W_1 \cdot P}, D \right)
\]

\textbf{Step 2: Replicate} the first dimension $N_1 \cdot P$ times and \textbf{interleave} 3D features to match 2D features:
\[
\text{Replicated shape} \gets \left( N, \frac{H}{H_1 \cdot P}, \frac{W}{W_1 \cdot P}, D \right)
\]

\textbf{Step 3: Reshape} interleaved features into $N$ slices:
\[
\{z_{\text{3D}}^j\} \gets \text{Reshape to} \ (N, L, D),
\]

\Return \{$z_{\text{3D}}^j$\}
\end{algorithm}

\begin{figure}[t]
\centerline{\includegraphics[width=1\linewidth]{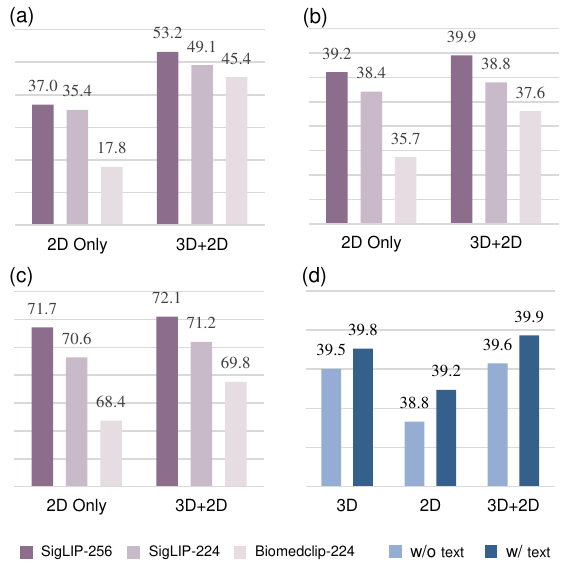}}
\caption{Ablation study.}
\label{fig4}
\end{figure}

\subsection{Ablation Study}
\label{sec:abl}

As shown in Fig.~\ref{fig4} (a) (b) (c), different 2D image encoders are evaluated in report generation (BLEU), open-ended VQA (BLEU), and closed-ended VQA (Accuracy) tasks. As expected, the model utilizing SigLIP at $256\times256$ outperforms the one with $224\times224$, emphasizing the importance of high image resolution. However, somewhat counterintuitively, Biomedclip \cite{biomedclip}, pre-trained on medical images at $224\times224$, exhibits inferior performance compared to SigLIP-224. This discrepancy may be attributed to two factors: first, Biomedclip was primarily pre-trained on 2D medical images, lacking sufficient understanding of 3D images; second, SigLIP was pre-trained on a large-scale dataset, providing it with strong general visual capabilities, enhanced generalization, and better transferability.

Additionally, given that different tasks require distinct inter-slice attention distributions for 3D medical images, the TG-IS scoring module is designed to incorporate both image features and text guidance. Based on the VQA task of M3D-Data, which includes a diverse range of question types, various combinations of image features, including 3D, 2D, and 3D+2D (Med-2E3), are compared, along with an evaluation of the impact of integrating text guidance.

As shown in Fig.~\ref{fig4} (d), different TG-IS scoring module designs are evaluated in the open-ended VQA task (BLEU). The TG-IS scoring module design adopted in Med-2E3, which integrates 3D+2D image features with text guidance, achieves the best performance. The results further indicate that utilizing 3D+2D features outperforms using 3D or 2D alone, highlighting the importance of feature completeness. Moreover, with the same image features, incorporating text guidance consistently enhances performance. These findings validate the effectiveness of the proposed TG-IS scoring module.

\begin{figure}[t]
\centerline{\includegraphics[width=1\linewidth]{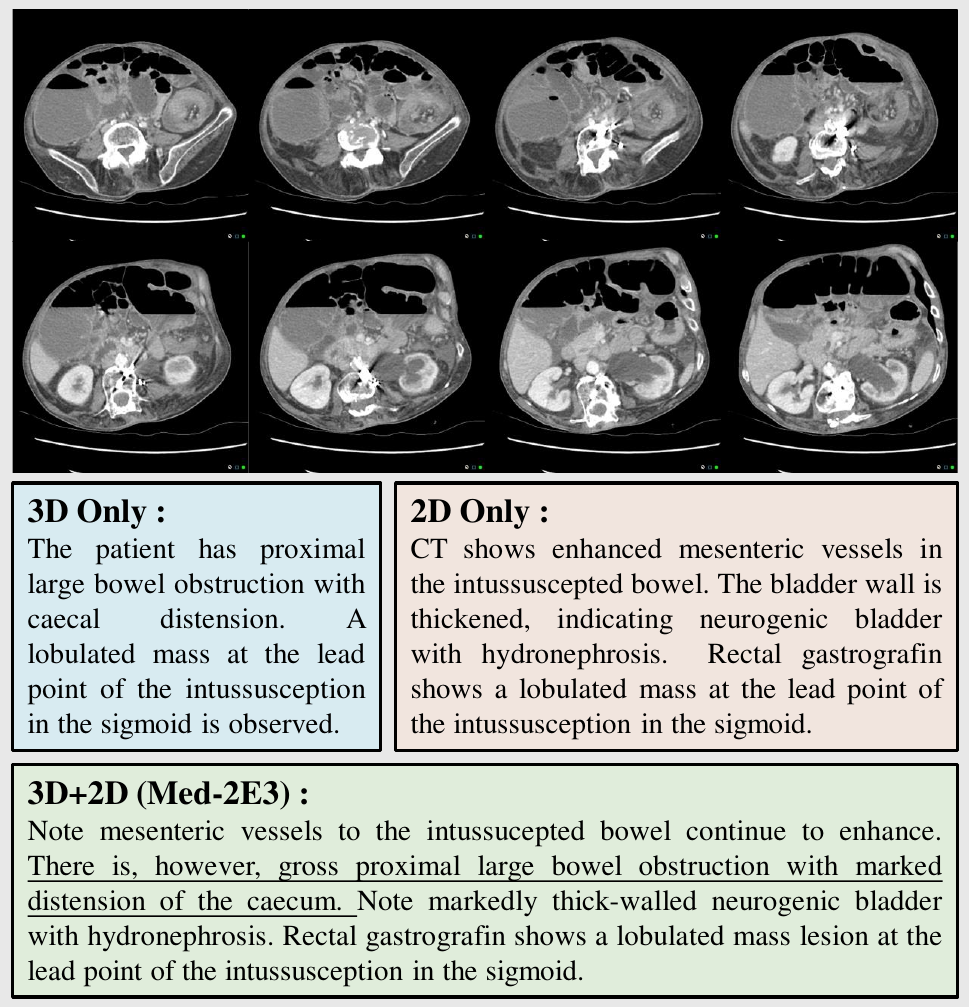}}
\caption{Reports generated by MLLMs using different image feature extraction paradigms, including single-encoder and dual 3D-2D encoder architectures. 3D+2D (Med-2E3) produces the most comprehensive results compared to 3D-only and 2D-only models. The 3D-only model lacks detail, while the 2D-only model omits certain overall observations, which are \underline{underlined}.}
\label{fig5}
\end{figure}

\subsection{Case Study}

As shown in Fig.~\ref{fig5}, a comparison of report generation results across different image feature extraction paradigms demonstrates that the 3D+2D model (Med-2E3) produces the most comprehensive and detailed interpretation. While the 3D-only model provides relevant observations, it lacks finer details present in the 3D+2D report. For example, it fails to mention the involvement of mesenteric vessels and the neurogenic bladder, resulting in a more generalized description that limits diagnostic depth. Similarity, the 2D-only model lacks broader contextual information for accessing disease progression. Although it identifies the thickened bladder wall and lobulated mass, it does not capture the overall severity of bowel obstruction or the enhancement of mesenteric vessels, both of which are crucial for a complete clinical assessment.

\end{document}